\documentclass[runningheads]{llncs}

\usepackage{graphicx}
\usepackage[caption=false]{subfig}
\usepackage{multirow}
\usepackage{color}
\usepackage{hyperref}

\begin{document}

\title{Learning Explainable Representations of Malware Behavior \thanks{This is a pre-print  of an article to appear in  Machine Learning
and Knowledge Discovery in Databases. ECML PKDD 2021.}}

%
%
%

\author{Paul Prasse\inst{1}\and
Jan Brabec\inst{2}\and
Jan Kohout\inst{2}\and\\
Martin Kopp\inst{2}\and
Lukas Bajer\inst{2}\and
Tobias Scheffer\inst{1}}

\authorrunning{P. Prasse et al.}
%
\institute{University of Potsdam, Department of Computer Science, Germany 
\email{\{prasse,scheffer\}@uni-potsdam.de}
\and
Cisco Systems, Cognitive Intelligence, Prague, Czech Republic \\
\email{\{janbrabe,jkohout,markopp,lubajer\}@cisco.com}}

\maketitle 
\begin{abstract}
We address the problems of identifying malware in network telemetry logs and providing \emph{indicators of compromise}---comprehensible explanations of behavioral patterns that identify the threat. In our system, an array of specialized detectors abstracts network-flow data into comprehensible \emph{network events} in a first step. We develop a neural network that processes this sequence of events and identifies specific threats, malware families and broad categories of malware. We then use the \emph{integrated-gradients} method to highlight events that jointly constitute the characteristic behavioral pattern of the threat. We compare network architectures based on CNNs, LSTMs, and transformers, and explore the efficacy of unsupervised pre-training experimentally on large-scale telemetry data. We demonstrate how this system detects njRAT and other malware based on behavioral patterns.


\end{abstract}

\section{Introduction}
\label{sec:introduction}

Toady's malware can exhibit different kinds of malicious behaviour. Malware collects personal and financial data, can encrypt users' files for ransom, is used to commit click-fraud, or promotes financial scams by intrusive advertising. Client-based antivirus tools employ  signature-based analysis, static analysis of portable-executable files, emulation,
and dynamic, behavior-based analysis to detect malware~\cite{swinnen2014one}. Systems that analyze network telemetry data complement antivirus software and are widely used in corporate networks. They allow organizations to enforce acceptable-use and security policies throughout the network and minimize management overhead. Telemetry analysis makes it possible to encapsulate malware detection into network devices or cloud services~\cite{karim2005malware,bartos2015robust}.

Research on applying machine learning to malware detection is abundant. However, the principal obstacle that impedes the deployment of machine-learning solutions in practice is that computer-security analysts need to be able to validate and confirm---or overturn---decisions to block software as malware.
However, machine-learning models usually work as black boxes and do not provide a decision rationale that analysts can understand and verify. In computer security, {\em indicators of compromise} refer to specific, observable evidence that indicates, with high confidence, malicious behavior. Security analysts consider indicators of compromise to be grounds for the classification of software as malware. For instance, indicators of compromise that identify software as variants of the \emph{WannaCry} malware family include the presence of the WannaCry ransom note in the executable file and communication patterns to specific URLs that are used exclusively by a kill-switch mechanism of the virus~\cite{wannacryiocs}.


In recent years, machine-learning models have been developed that emphasize \emph{explainability} of the decisions and underlying representations. For instance, \emph{Shapley values}~\cite{shap}, and the  \emph{DeepLift}~\cite{shrikumar2016not} and \emph{integrated gradients} methods~\cite{sundararajan2017axiomatic} quantify the contribution of input attributes to the model decision. However, in order to be part of a comprehensible explanation of why software is in fact malicious, the importance weights would have to refer to events that analysts can relate to specific behavior of malicious software. 

In this paper, we first discuss a framework of classifiers that detect a wide range of intuitively meaningful network events. We then develop neural networks that detect malware based on behavioral patterns composed of these behaviors. We compare network architectures based on CNNs, LSTMs, and transformers. In order to address the relative scarcity of labeled data, we investigate whether initializing the models by unsupervised pre-training improves their performance. We review how the model detects the njRAT and other malware families based on behavioral indicators of compromise.



\section{Related Work}
\label{sec:related}

Prior work on the analysis of {\em HTTP logs}~\cite{nguyen2008survey} has addressed the problems of identifying command-and-control servers~\cite{nelms2013execscent}, unsupervised detection of malware~\cite{kohout2015unsupervised,bartos2016optimized}, and supervised detection of malware using domain blacklists as labels~\cite{franc2015learning,bartos2015robust,bayesianforests}.
HTTP log files contain the full URL string, from which a wide array of informative features can be extracted~\cite{bartos2015robust}. 

A body of recent work has aimed at detecting Android malware by network-traffic analysis. Arora \emph{et al.}~\cite{arora2014malware} use the average packet size, average flow duration, and a small set of other features to identify 48 malicious Android apps. Lashkari \emph{et al.}~\cite{lashkari2015towards} collect 1,500 benign and 400 malicious Android apps, extract flow duration and volume feature, and apply several machine-learning algorithms from the Weka library. They observe high accuracy values on the level of individual flows. 
Demontie \emph{et al.}~\cite{demontis2017yes} model different types of attacks against such detection mechanisms and devise a feature-learning paradigm that mitigates these attacks.
Malik and Kaushal~\cite{malik2016credroid} aggregate the VirusTotal ranking of an app with a crowd-sourced domain-reputation service (Web of Trust) and the app's resource permission to arrive at a ranking. 

Prior work on {\em HTTPS logs} has aimed at identifying the application layer protocol~\cite{wright2006on,crotti2007traffic,dusi2009tunnel}. In order to cluster web servers that host similar applications, Kohout \emph{et al.}~\cite{kohout2015automatic} developed features that are derived from a histogram of observable time intervals and data volumes of connections. Using this feature representation, Loko{\v{c}} \emph{et al.}~\cite{lokovc2016k} introduced an approximate $k$-NN classifier that identifies servers which are contacted by malware. 

Graph-based classification methods~\cite{anderson2011graph} have been explored, but they can only be applied by an agent that is able to perceive a significant portion of the global network graph---which raises substantial logistic and privacy challenges. By contrast, this paper studies an approach that relies only on the agent's ability to observe the traffic of a single organization.

Prior work on neural networks for network-flow analysis~\cite{pevny2016discriminative} has worked with labels for client computers (infected and not infected)---which leads to a multi-instance learning problem. 
CNNs have also been applied to analyzing URLs which are observable as long as clients use the HTTP instead of the encrypted HTTPS protocol~\cite{saxe2017expose}.
Malware detection from HTTPS traffic has been studied using a combination of word2vec embeddings of domain names and long short term memory networks (LSTMs)~\cite{prasse2017ecml} as well as convolutional neural networks~\cite{prasse2019joint}. Since the network-flow data only logs communication events between clients and hosts, these models act as black boxes that do not provide security analysts any verifiable decision rationale. Since we collected data containing only specific network events without the information of the used domain names, we are not able to apply these models to our data.

Recent findings suggest that the greater robustness of convolutional neural networks (CNNs) may outweight the ability of LSTMs to account for long-term dependencies~\cite{gehring2017convolutional}. This motivates us to explore convolutional architectures. Transformer networks~\cite{vaswani2017attention} are encoder-decoder architectures using multi-head self-attention layers and positional encodings widely used for NLP tasks. GPT-2~\cite{radford2019better} and BERT~\cite{devlin2018bert} show that transformers pre-trained on a large corpus learn representations that can be fine-tuned for classification problems.

\section{Problem Setting and Operating Environment}
\label{sec:data_collection}

This section first describes the operating environment and the first stage of the \emph{Cisco Cognitive Intelligence} system that abstracts network traffic into \emph{network events}. Section~\ref{sec:threats} proceeds to define the threat taxonomy and to lay out the problem setting. Section~\ref{sec:data} describes the data set that we collect for the experiments described in this paper. 

\subsection{Network Events}
\label{sec:network_events}

The \emph{Cisco Cognitive Intelligence (CI)}~\cite{cognitiveblog} intrusion detection system monitors the network traffic of the customer organization for which it is deployed. Initially, the traffic is captured in the form of web proxy logs that enumerate which users connect to which servers on the internet, and include timestamps and the data volume sent and received. The CI engine then abstracts log entries into a set of \emph{network events}---high-level behavioral indicators that can be interpreted by security analysts. Individual network events are not generally suspicious by themselves, but specific \emph{patterns of network events} can constitute \emph{indicators of compromise} that identify threats. 
%
In total, CI distinguishes hundreds of events; their detection mechanisms fall into four main categories.

\begin{itemize}
\item \emph{Signature-based events} are detected by matching behavioral signatures that have been created manually by a domain expert. This includes detection based on known URL patterns or known host names.
\item \emph{Classifier-based events} are detected by special-purpose classifiers that have been trained on historical proxy logs. These classifiers included models that identify specific popular applications.
\item \emph{Anomaly-based events} are detected by a multitude of statistical, volumetric, proximity-based, targeted, and domain-specific anomaly detectors. Events in this category include, for  example, contacting a  server which is unlikely for the given user, or communication patterns that are too regular to be caused by a human user using a web browser.
\item \emph{Contextual events} capture various network behaviors to provide additional context; for instance, file downloads, direct access of a raw IP address without specified  host name, or software updates.
\end{itemize}
For purposes of the work, each interval of five minutes in which at least one network flow is detected, the \emph{set of network events} is timestamped and logged. Events are indexed by the \emph{users} who sent or received the traffic. No data are logged for intervals in which no event occurs. The resulting data structure for each organization is a sparse sequence of sets of network events for each user within the organization. 

\subsection{Identification of Threats}\label{sec:threats}

We use a malware taxonomy with three levels: \textit{threat ID, malware family, and malware category}. The \emph{threat ID} identifies a particular version of a malware product, or versions that are so similar that a security analyst cannot distinguish them. For instance, a threat ID can correspond to a particular version of the njRAT malware~\cite{njrat}, all instances of which use the same user-agent and URL pattern for communication. The \emph{malware family} entails all versions of a malware product---for instance, WannaCry is a malware family of which multiple versions are known to differ in their communication behavior. Finally, the \emph{malware category} broadly characterizes the monetization scheme or harmful behavior of a wide range of malware products. For instance, \emph{advertisement injector}, \emph{information stealer}, and \emph{cryptocurrency miner} are malware categories. 

Labeled training and evaluation data consist of sets of network events of five-minute intervals associated with a particular user in which threats have been identified by security analysts. In order to determine threat IDs, malware families, and categories, security analysts inspect network events and any available external sources of information about contacted servers. In some cases, hash keys of the executable files are also available and can be matched against databases of known malware to determine the ground truth. Due to this involvement of qualified experts, labeled data are valuable and relatively scarce. 

The \emph{problem setting} for the malware-detection model is to detect for each organization, user, and each five-minute interval in which at least one network event has occurred, which threat ID, malware family, and malware category the user has been exposed to. That is, each \emph{instance} is a combination of an organization, a user and a five-minute time interval.
Threats are presented to security analysts on the most specific level on which they can be detected. Specific threat IDs provide the most concrete actionable information for analysts. However, for unknown or unidentifiable threats, the malware family or category provides a lead which an analyst can follow up on. In addition to the threat, \emph{indicators of compromise} in the form of the relevant network events that identify the threat have to be  presented to the analysts for review. 

The analysis of this paper focuses on distinguishing between different threat IDs, malware families, and categories, and offering comprehensive indicators of compromise. The equally important problem of distinguishing between malware and benign activities has, for instance, been studied by Prasse \emph{et al.}~\cite{prasse2019joint}.
The majority of benign network traffic is not included in our data because only time intervals in which network events occur are logged. 

We will measure precision-recall curves, the multi-class accuracy, and the macro-averaged AUC to evaluate the models under investigation. The average AUC is calculated as the mean of the AUC values of the individual classes. Precision---the fraction of alarms that are not false alarms---directly measures the amount of unnecessary workload imposed on security analysts, while recall quantifies the detection rate. We also compare the models in terms of ROC cuves because these curves are invariant to class ratios.



\begin{table}[t]
	\caption{Data set statistics for malware category evaluation.}
	{\small\begin{center}
		\begin{tabular}
			{l|r|r}
			Malware category &Training instances& Test instances  \\ \hline\hline
            Potentially unwanted application & 14,675 & 10,026 \\ \hline
            Ad injector & 14,434 & 17,174 \\ \hline
            Malicious advertising & 3,287 & 1,354 \\ \hline
            Malicious content distribution & 2,232 & 9,088 \\ \hline
            Cryptocurrency miner & 1,114 & 1,857 \\ \hline
            Scareware & 198 & 398 \\ \hline
            Information stealer & 128 & 131
		\end{tabular}
	\end{center}}
\label{tab:data_malware_category}
\end{table}

\subsection{Data Collection and Quantitative Analysis}\label{sec:data}

We collected the entire network traffic of 348 companies for one day in June 2020 as training data, and for one day in July 2020 as evaluation data. The training data contain the network traffic of 1,506,105 users while the evaluation data contain the traffic of 1,402,554 unique users. 
In total, the data set consists of 9,776,911 training instances and 9,970,560 test instances, where each instance is a combination of an organization, a user, and a five-minute interval in which at least one network event was observed. In total, 216 distinct network events occur at least once in training and evaluation data---most of these events occur frequently. On average, 2.69 network events are observed in each five-minute interval in the training data and 2.73 events in the test data. 


Table~\ref{tab:data_malware_category} shows the seven malware categories that occur in the data at least 100 times. \emph{Potentially unwanted applications (PUAs)} are the most frequent class of malware; these free applications are mostly installed voluntarily for some advertised functionality, but then cannot be uninstalled without expert knowledge and expose the user to intrusive advertisements or steal user data. 
Table~\ref{tab:data_malware_fam} shows all malware families that analysts have identified in our data. Most malware families fall into the category of PUA, but analysts have been able to identify a number of high-risk viruses. Comparing Tables~\ref{tab:data_malware_category} and ~\ref{tab:data_malware_fam} shows that for many threats, analysts are able to determine the malware category, but not the specific malware family. 

Finally, Table~\ref{tab:data_cluster_label} shows those threat IDs for which at least 100 instances occur in our data. In many cases, analysts identify a specific threat which is assigned a threat ID based on the malware's behavior, without being able to ultimately determine which malware family it has been derived from.

\begin{table}[t]
	\caption{Data set statistics for malware family evaluation.}
	{\small\begin{center}
		\begin{tabular}
			{l|r|r|l}
			Malware family &Training & Test & Malware category  \\ \hline\hline
			ArcadeYum & 12,051 & 6,231 & Potentially unwanted application \\ \hline
			Malicious Android firmware & 38 & 30 & Information stealer \\ \hline
            njRAT & 15 & 37 & Information stealer \\ \hline
            WannaCry & 4 & 7 &  Ransomware \\ \hline
		\end{tabular}
	\end{center}}
\label{tab:data_malware_fam}
\end{table}


\begin{table}[t!]
	\caption{Data set statistics for threat ID evaluation.}
	{\small\begin{center}
		\begin{tabular}
			{l|r|r|l}
			Threat ID &Training instances& Test instances& Malware category  \\ \hline\hline 
            Threat ID 1
            & 8,900 & 9,710 & Ad injector \\ \hline
            Threat ID 2
            & 900 & 924 & Potentially unwanted application\\ \hline
            Threat ID 3
            & 11,894 & 6,075 & Potentially unwanted application \\ \hline
            Threat ID 4
            & 641 & 783 & Potentially unwanted application \\ \hline
            Threat ID 5
            & 606 & 425 & Ad injector \\ \hline
            Threat ID 6
            & 392 & 567 & Malicious advertising \\ \hline
            Threat ID 7
            & 2,099 & 9,027 & Malicious content distribution \\ \hline
            Threat ID 8
            & 119 & 54 &  Typosquatting\\ \hline
            Threat ID 9
            & 282 & 193 &  Phishing
		\end{tabular}
	\end{center}}
\label{tab:data_cluster_label}
\end{table}

\section{Models}
\label{sec:model}

\begin{figure}[t!]
	\centering
	 \includegraphics[width=\linewidth,keepaspectratio,
	 clip]{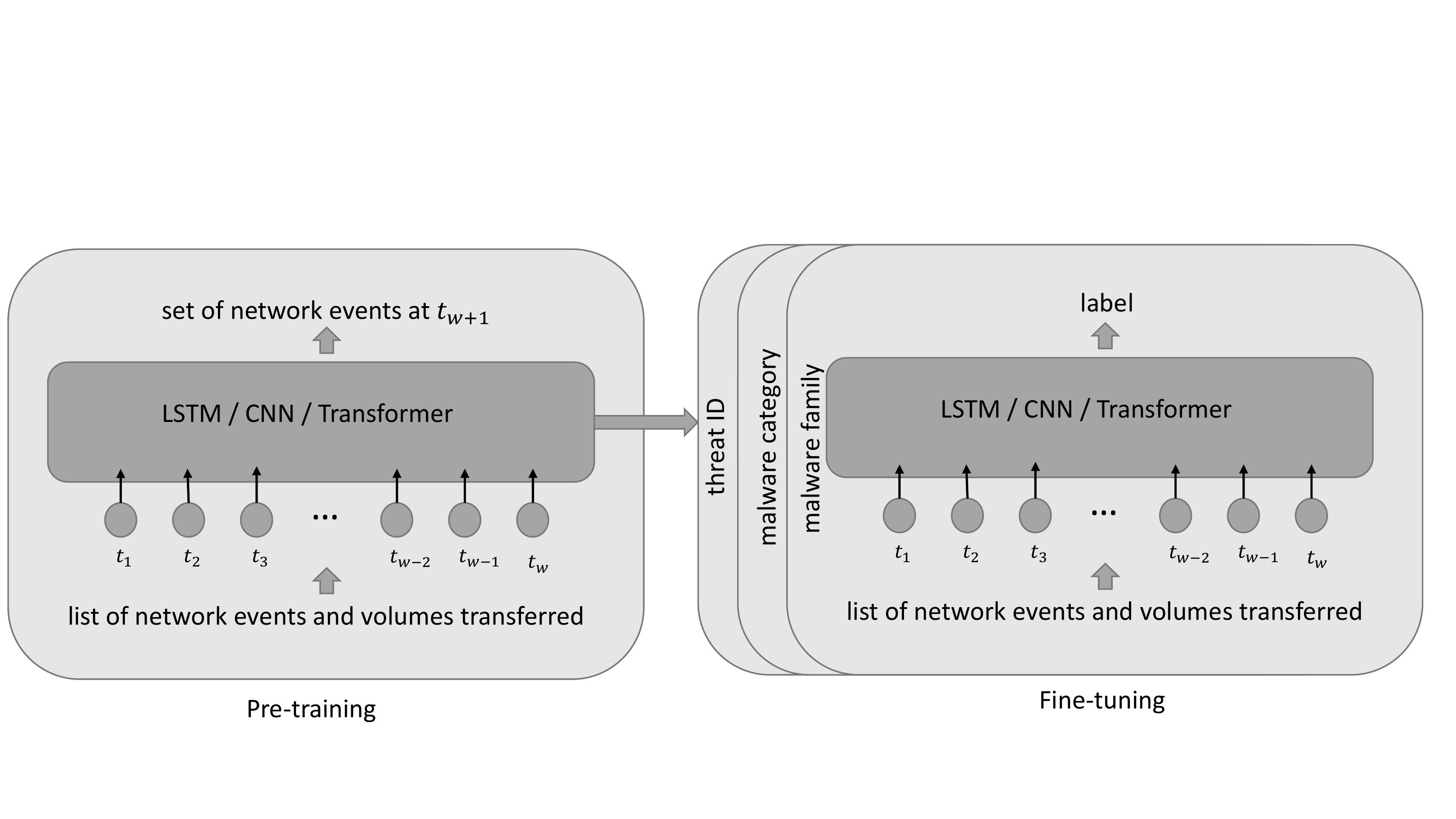}
		\caption{Model architecture.}
	\label{fig:architecture_overview}
\end{figure}

\begin{figure}[t!]
	\centering
	\subfloat[Model input for a single five-minute interval.\label{fig:model_input}]{
	 \includegraphics[width=0.40\linewidth,keepaspectratio,trim=0 0 550 200,clip]{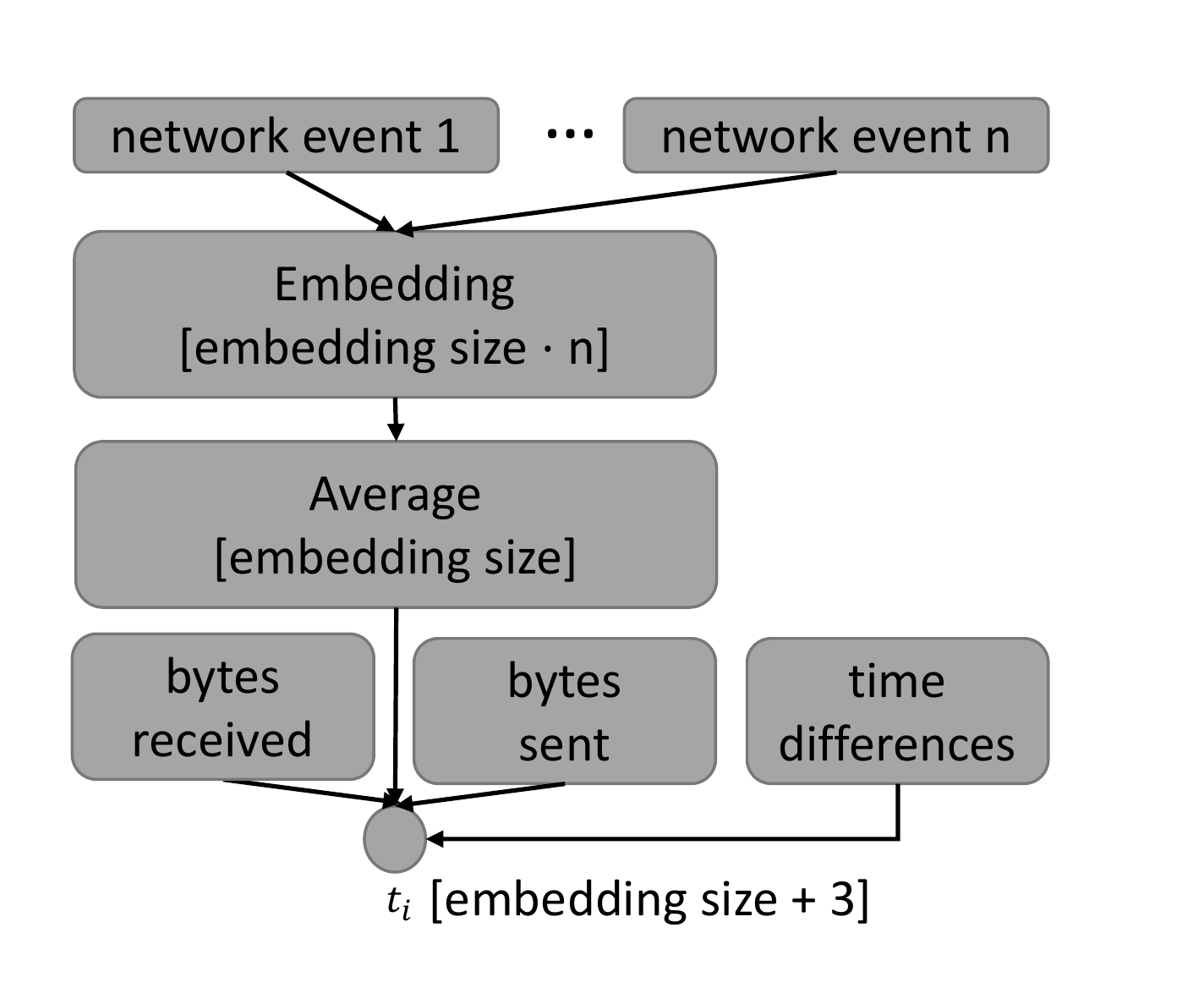}}
	 \hfil
    \subfloat[LSTM model architecture.\label{fig:model_lstm}]{
        \includegraphics[width=0.49\linewidth,keepaspectratio,trim=25 15 550 200,clip]{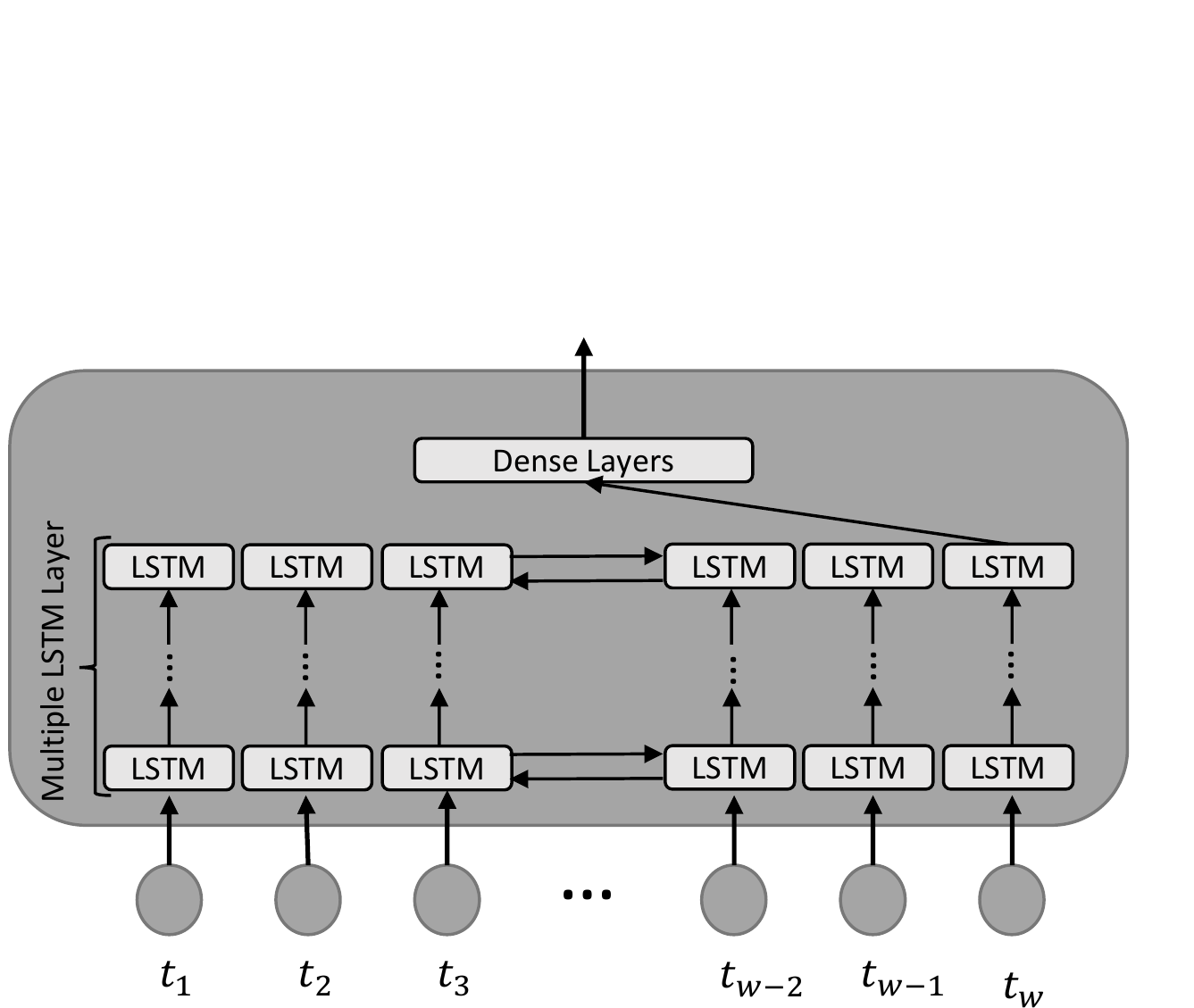}}
    \hfil
    \hfil
    \subfloat[CNN model architecture.\label{fig:model_cnn}]{
        \includegraphics[width=0.49\linewidth,keepaspectratio,trim=25 15 550 200,clip]{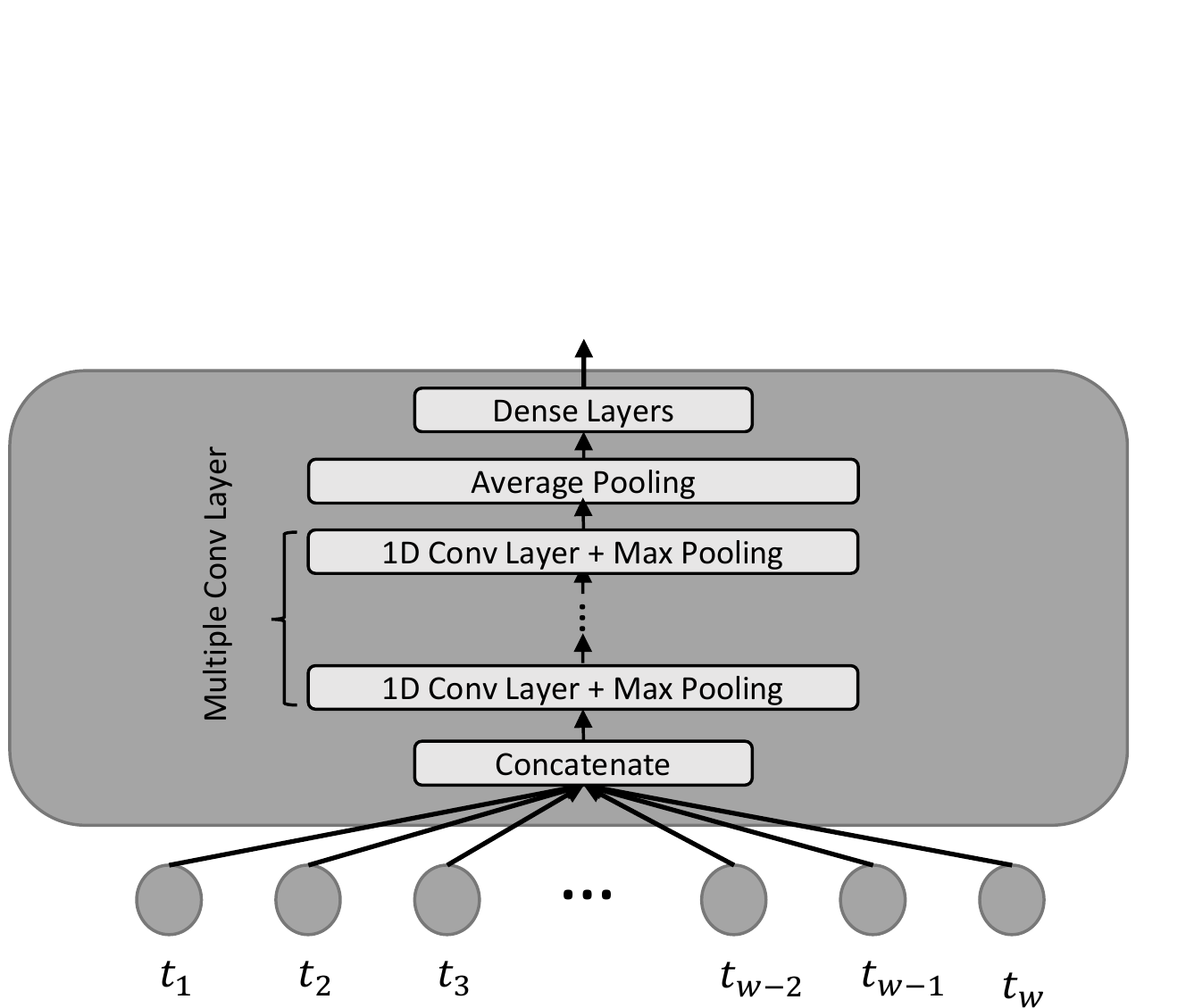}}
    \hfil
    \hfil
    \subfloat[Transformer model architecture.\label{fig:model_transformer}]{
        \includegraphics[width=0.49\linewidth,keepaspectratio,trim=25 15 550 200,clip]{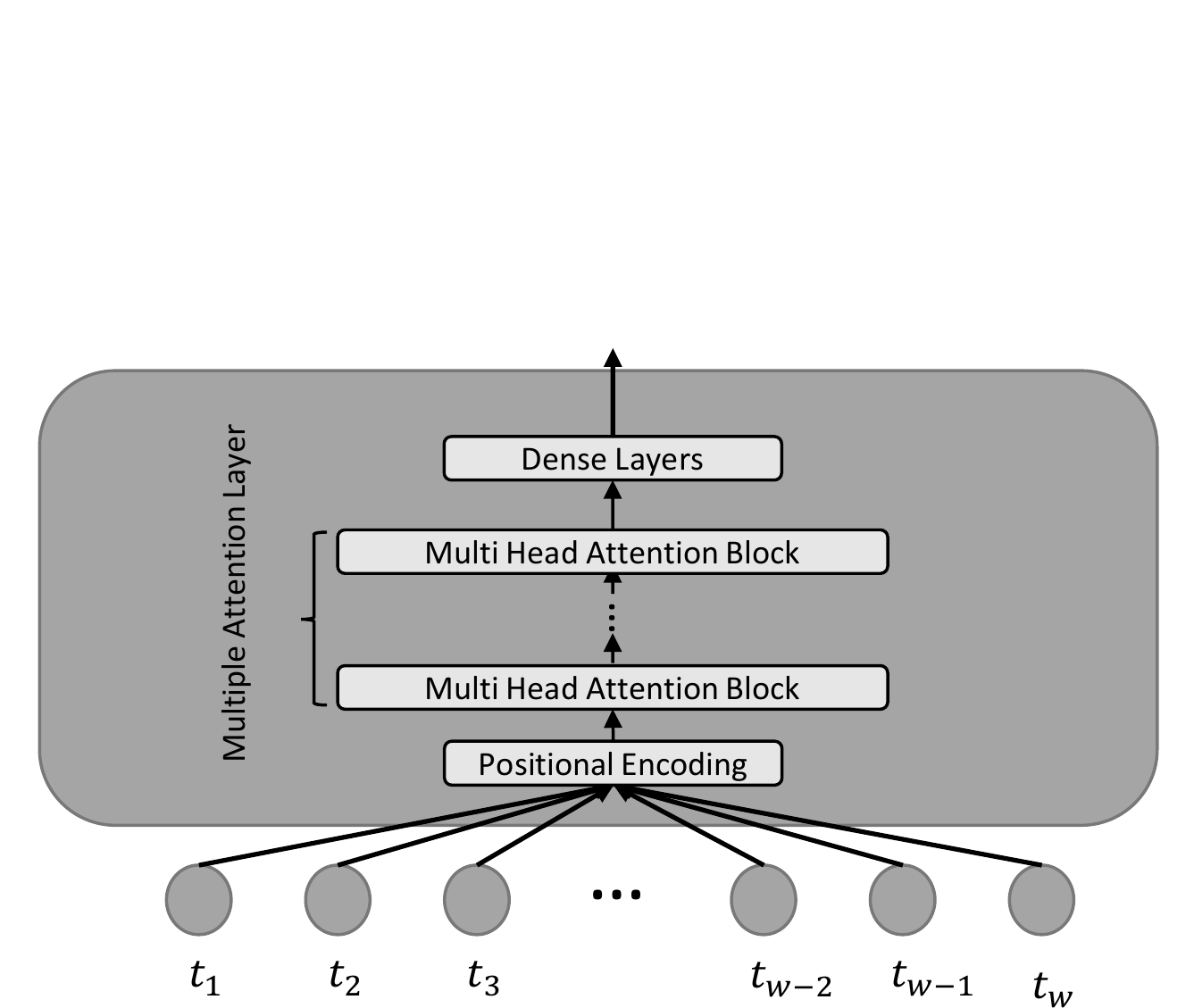}}
	\caption{Model input and models.}
	\label{fig:models}
\end{figure}

This section develops the neural network architectures that we will explore in the experimental part of this paper. All networks process the sequence of network events provided by the detector array. In one version of the networks, we employ unsupervised pre-training of the models---see Figure \ref{fig:architecture_overview}. We will compare the pre-trained models to reference versions without pre-training. 

\subsection{Architectures} \label{sec:architectures}

Here we develop three different model architectures: an \textit{LSTM} model using severeral bidirectional LSTM layers~\cite{hochreiter1997long}, a \textit{CNN} model using stacked one-dimensional CNN layers~\cite{gehring2017convolutional}, and a \textit{transformer} model that uses multiple multi-head attention blocks~\cite{vaswani2017attention}. We also implement a random forest baseline.

The input to the different model architectures consists of a window of $w$ five-minute intervals, each of which is represented by a set of network events, a timestamp, and the numbers of bytes sent and received. The width $w$ of the window is a tunable hyperparameter. The set of network events for each time step are processed by an embedding layer followed by an averaging layer that computes the mean embedding for all the network events for the current five-minute interval (see Figure~\ref{fig:model_input}). The mean embedding is than concatenated with the log transformed time differences between subsequent elements in the window and the log-transformed number of bytes sent and received.

The LSTM model consists of multiple layers of bidirectional LSTM units, followed by a number of dense layers with a dropout rate of 0.1. The number of layers of each type and the number of units per layer for each of the models are hyperparameters that we will tune in Section~\ref{sec:experiments}; see Table~\ref{tab:best_parameter}. The output layer consists of a softmax layer with the number of units equal to the number of different classes (see Figure \ref{fig:model_lstm}).
The CNN model starts off with a concatenation layer that combines the elements in the input window. The next layers are multiple pairs of a one-dimensional convolutional layer followed by a max-pooling layer. The last CNN layer is connected to an average-pooling layer and a number of dense layers on top of it. The last layer is a softmax layer with one unit per output classes (see Figure \ref{fig:model_cnn}).

The transformer model consists of an absolute positional encoding layer that outputs the sum of the positional encoding and the concatenated input sequence~\cite{shaw2018self}. The output of the positional encoding layer is fed into multiple attention layers~\cite{vaswani2017attention}. The output of the last multi-head attention layer is fed into a sequence of dense layers. The last layer consists of a softmax layer with one unit per output class (see Figure \ref{fig:model_transformer}).

The random forest (\textit{RF}), which serves as natural baseline, consumes the one-hot encodings of all network events within the window and the concatenated list of all log transformed bytes sent, log-transformed bytes received, and the time differences between susequent elements within the window.

\subsection{Unsupervised Pre-training} \label{sec:pre-training}

Since labeling training data requires highly-trained analysts to identify and analyze threats, labeled data are relatively scarce. While the number of labels is in the tens of thousands in our study, the number of unlabeled instances collected over two days is around 20 millions. Unsupervised pre-training offers the potential for the network to learn a low-level representation that structures the data in such a way that the subsequent supervised learning problem becomes easier to solve with limited labeled data.

To pre-train the models, we use all 9,776,911 training instances. The training objective is to predict the set of network events present at time step $t_{w+1}$ given the sets of events of previous time steps $t_1,\dots t_w$ (see Figure \ref{fig:architecture_overview}). This is a multi-label classification problem, since we want to predict all present network events at time step $t_{w+1}$. This model serves as a ``language model''~\cite{radford2019better,devlin2018bert} for network events that learns an internal representation which is able to predict the next network events given their context. 
For the pre-training step, we add a fully connected dense layer with sigmoid activation function to the models. We train these models using the binary cross entropy loss function. We will compare the \textit{pre-trained} models to their counterparts that have been trained  \textit{from scratch} with Glorot initialization~\cite{pmlr-v9-glorot10a}.

\section{Experiments}
\label{sec:experiments}

\begin{table}[t!]
\begin{center}
\caption{Best hyperparameters found using grid search.}
{\scriptsize\begin{tabular}{cl|c||l}
&{hyperparameter} &  {parameter range} & {best value} \\ \hline
\multirow{5}{*}{\rotatebox[origin=c]{90}{LSTM}} & embedding size & $\{2^{6},\dots,2^{8}\}$  & 128\\ 
& \# LSTM layers & $\{1,\cdots,4\}$ & 1 \\
& LSTM units & $\{2^3,\dots,2^{11}\}$ & 1024 \\
& \# Dense layers & $\{1,\dots,3\}$ & 2 \\
& \# Dense units & $\{2^6,\dots, 2^{10}\}$ & 256 \\ \hline
\multirow{5}{*}{\rotatebox[origin=c]{90}{CNN}} & embedding size & $\{2^{6},\dots,2^{8}\}$  & 128\\ 
& \# CNN layers & $\{1,\cdots,4\}$ & 3 \\
& kernel size & $\{2^1,\dots,2^{3}\}$ & 4 \\
& \# filters & $\{2^2,\dots,2^{7}\}$ & 32 \\
& \# Dense layers & $\{1,\dots,3\}$ & 2 \\
& \# Dense units & $\{2^6,\dots, 2^{10}\}$ & 256 \\ \hline
\multirow{6}{*}{\rotatebox[origin=c]{90}{Transformer}} & embedding size & $\{2^{6},\dots,2^{8}\}$  & 128\\ 
& \# attention blocks & $\{1,\cdots,4\}$ & 2 \\
& \# attention heads & $\{2^2,\dots,2^{7}\}$ & 8 \\
& \# Dense attention units & $\{2^2,\dots,2^{7}\}$ & 512 \\
& \# Dense layers & $\{1,\dots,3\}$ & 2 \\
& \# Dense units & $\{2^6,\dots, 2^{10}\}$ & 512 \\ \hline
\multirow{2}{*}{\rotatebox[origin=c]{90}{RF}} & \# trees & $\{10,100,1000\}$  & 100\\ 
& \# max depth & $\{2,10,100,None\}$ & 10 \\
\label{tab:best_parameter}
\end{tabular}}
\end{center}
\end{table}

This section reports on malware-detection performance of the models under investigation, and on the interpretability of the indicators of compromise. We split the data into a training part that we acquired in June 2020 and an evaluation part acquired in July 2020.

\subsection{Hyperparameter Optimization} \label{sec:param_opt} 

We optimize the width of the window of five-minute time intervals $w$ used to train the models by evaluating values from 3 to 41 with a nested training-test split on the training part of the data using the threat-ID classification task. In the following experiments, we fix the number of used five-minute intervals $w$ to 21 (see Figure~\ref{fig:models}). That is, each training and test instance is a sequence of 21 five-minute intervals; training and test sequences are split into overlapping sequences of that length. 
We tune the number of layers of each type, and the number of units per layer for all models using a 5-fold cross-validation on the training part of the data using the threat-ID classification task. The grid of parameters and the best hyperparameters can be found in Table \ref{tab:best_parameter}. The optimal parameters for the random forest baseline are found using a 5-fold cross-validation on the training data of the given task.

We train all models on a single server with 40-core Intel(R) Xeon(R) CPU E5-2640 processor and 512 GB of memory. We train all neural networks using the Keras
and Tensorflow
libraries on a GeForce GTX TITAN X GPU using the NVidia CUDA platform. We implement the evaluation framework using the scikit-learn
machine learning package. The code can be found online\footnote{\scriptsize{\url{https://github.com/prassepaul/Learning-Explainable-Representations-of-Malware-Behavior}}}.

\subsection{Malware-Classification Performance} \label{sec:performance}


In the following we compare the classification performance of the different models for the tasks of detecting threat IDs, malware categories, and malware families. We compare neural networks that are trained from scratch using Glorot initialization and models initialized with pre-trained weights as described in Section~\ref{sec:pre-training}. 
We also investigate how the number of training data points per class effects the performance. To do so, we measure the accuracy acc@$n$ and average AUC@$n$ after the models have been trained on $n$ instances per class. Since obtaining malware labels is time consuming and costly, this gives us an estimation of how the models behave in a few-shot learning scenario.

Table~\ref{tab:malware_types} shows the overall results for all described models and all the different levels of the threat taxonomy on the evaluation data. We see that the transformer outperforms CNN and LSTM most of the time, and that the pre-trained models almost always significantly outperform their counterparts that have been trained from scratch, based on a two-sided, paired $t$ test with $p<0.05$. Only the LSTM models are in some cases not able to benefit from  pre-training. We also see that the neural network architectures outperform the random forest baseline in all settings, so we conclude that using the sequential information and sequential patterns can be exploited to classify different malware types. Using more training instances nearly always boosts the overall performance. Only for the detection of different malware families the performance in terms of the average AUC is lower when training with the full data set. We think this is caused by highly imbalanced class distribution pushing the models to favor for specific classes.

From Table~\ref{tab:malware_types}, we conclude that in almost all cases the transformer model with unsupervised pre-training is the overall best model. Because of that, the following detailed analysis is performed using only the transformer model architecture.  

Additional experiments in which we determine the ROC and precision-recall curves that the transformer with pre-training achieves for individual threats, malware families, and malware categories can be found in Appendix \ref{sec:appendix}. From these experiments, we can furthermore conclude that threat IDs that have a one-to-one relationship with a malware family are the easiest ones to identify, and that broad categories such as PUA that include a wide range of different threats are the most difficult to pin down.

\begin{table}[ht!]
	\caption{Accuracy and AUC for the detection of threat IDs, malware families, and categories, after training on some or all training data, with and without pre-training. Acc@$n$ and AUC@$n$ refer to the accuracy and AUC, respectively, after training on up to $n$ instances per class. For results marked ``*'', the accuracy of pre-trained models is significantly better ($p<0.05$) compared to the same model trained from scratch. Results marked ``$\dagger$'' are significantly better ($p<0.05$) than the next-best model.}
	{\small\begin{center}
		\begin{tabular}
		{ll|c|c|c|c|c|c|c}
        & & \multicolumn{2}{c|}{CNN}& \multicolumn{2}{c|}{LSTM}& \multicolumn{2}{c|}{Transformer}& \multicolumn{1}{c}{Random Forest} \\ 
        & & Scratch& Pre-tr.& Scratch& Pre-tr.& Scratch& Pre-tr.& Scratch\\ \hline \hline
        \multirow{8}{*}{\rotatebox[origin=c]{90}{threat ID}} &acc@10 & 0.394 & 0.437* & 0.314 & 0.375* & 0.352 & \textbf{0.559*$\dagger$} & 0.413\\
        & acc@50 & 0.618 & 0.648* & 0.567 & 0.478 & 0.612 & \textbf{0.731*$\dagger$} & 0.57\\
        & acc@100 & 0.666 & 0.689* & 0.624 & 0.54 & 0.685 & \textbf{0.759*$\dagger$} & 0.614\\
        & acc & 0.785 & 0.799 & 0.806 & \textbf{0.843*$\dagger$} & 0.769 & 0.776 & 0.809\\ \cline{2-9}
        & AUC@10 & 0.794 & 0.773 & 0.75 & 0.698 & 0.748 & \textbf{0.848*} & 0.832\\
        & AUC@50 & 0.889 & 0.893 & 0.874 & 0.807 & 0.893 & \textbf{0.941*$\dagger$} & 0.902\\
        & AUC@100 & 0.906 & 0.914* & 0.897 & 0.829 & 0.902 & \textbf{0.948*$\dagger$} & 0.912\\
        & AUC & 0.937 & \textbf{0.952*} & 0.935 & 0.942 & 0.915 & 0.95* & 0.925\\ \hline \hline
        \multirow{8}{*}{\rotatebox[origin=c]{90}{malware category}} &acc@10 & 0.23 & 0.396* & 0.196 & 0.312* & 0.228 & \textbf{0.456*$\dagger$} & 0.338\\
        & acc@50 & 0.524 & 0.598* & 0.515 & 0.485 & 0.575 & \textbf{0.652*$\dagger$} & 0.54\\
        & acc@100 & 0.618 & 0.669* & 0.597 & 0.539 & 0.652 & \textbf{0.703*$\dagger$} & 0.606\\
        & acc & 0.77 & 0.785* & 0.769 & 0.772 & 0.771 & \textbf{0.802*$\dagger$} & 0.73\\ \cline{2-9}
        & AUC@10 & 0.752 & 0.813* & 0.73 & 0.728 & 0.747 & \textbf{0.821*} & 0.819\\
        & AUC@50 & 0.86 & 0.901* & 0.861 & 0.808 & 0.879 & \textbf{0.924*$\dagger$} & 0.889\\
        & AUC@100 & 0.881 & 0.91* & 0.877 & 0.831 & 0.914 & \textbf{0.938*$\dagger$} & 0.907\\
        & AUC & 0.917 & 0.937* & 0.908 & 0.902 & 0.912 & \textbf{0.96*$\dagger$} & 0.916\\ \hline \hline
        \multirow{8}{*}{\rotatebox[origin=c]{90}{malware family}} & acc@10 & 0.439 & \textbf{0.91*} & 0.01 & 0.894* & 0.322 & 0.893* & 0.846\\
        & acc@50 & 0.839 & 0.939* & 0.592 & 0.923* & 0.808 & \textbf{0.946*} & 0.923\\
        & acc@100 & 0.87 & 0.959 & 0.875 & 0.952 & 0.866 & \textbf{0.977*$\dagger$} & 0.962\\
        & acc & 0.929 & 0.993* & 0.889 & \textbf{0.995*$\dagger$} & 0.954 & 0.992 & 0.994\\ \cline{2-9}
        & AUC@10 & 0.886 & \textbf{0.985*} & 0.785 & 0.964* & 0.855 & 0.983* & 0.106\\
        & AUC@50 & 0.96 & 0.992* & 0.832 & 0.982* & 0.967 & \textbf{0.993*} & 0.199\\
        & AUC@100 & 0.96 & 0.993* & 0.922 & 0.983 & 0.969 & \textbf{0.995*} & 0.199\\
        & AUC & 0.921 & \textbf{0.983*} & 0.923 & \textbf{0.983*} & 0.486 & 0.947* & 0.322
		\end{tabular}
	\end{center}}
\label{tab:malware_types}
\end{table}

\subsection{Indicators of Compromise}
\label{sec:explanation}
This section explores the interpretablility of the indicators of compromise inferred from the transformer model. We use the \emph{integrated gradients} method to highlight the most important features for a given input sequence~\cite{sundararajan2017axiomatic}. Integrated gradients can compute the contribution of each network event when classifying a given input sequence. We calculate the impact of all input features using

\begin{equation}
    IG_{i}(x) = (x_{i} - x'_{i})\times\int_{\alpha=0}^1\frac{\partial F(x'+\alpha \times (x - x'))}{\partial x_i}{d\alpha},
    \label{eq:ig}
\end{equation}
where $_{i}$ denotes the $i$-th feature, $x$ the input to the model, $x'$ the baseline, and $\alpha$ the interpolation constant for the perturbation of the features. 
The term $(x_i - x'_i)$ denotes the difference between original input and ``baseline''. Similar to the all-zeros baseline that is used for input images, we set the baseline to the instance with all zero-embeddings and original numerical features. The baseline input is needed to scale the integrated gradients. In practice we  approximate this integral by the numerical approximation
\begin{equation}
    IG^{approx}_{i}(x) =(x_{i}-x'_{i})\times\sum_{k=1}^{m}\frac{\partial F(x' + \frac{k}{m}\times(x - x'))}{\partial x_{i}} \times \frac{1}{m}
    \label{eq:ig_approx},
\end{equation}
where $k$ is the number of approximation steps. 

\subsubsection{Single-Instance Evaluation} 

\begin{figure}[ht!]
	\centering
	 \includegraphics[width=1\linewidth,keepaspectratio, trim=40 150 55 0,clip]{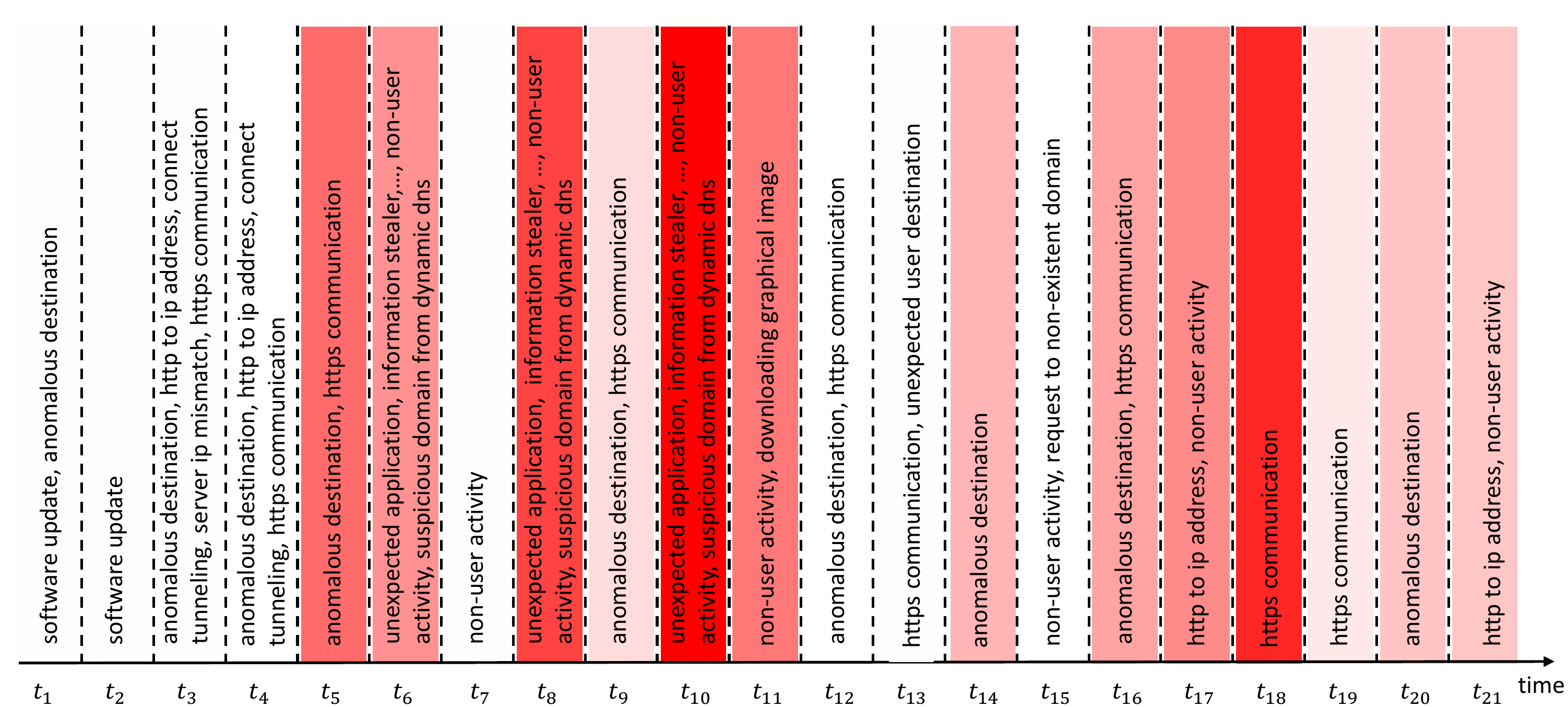}
		\caption{Feature importance for detection of njRAT using \emph{integrated gradients} for a single instance. The intensity of the red hue indicates the importance of  network events.}
	\label{fig:njrat}
\end{figure}

Using the Integrated Gradients from Equation~\ref{eq:ig_approx}, we determine which input time steps contributed to which extend to the overall classification. 
Figure~\ref{fig:njrat} shows an example output for an instance classified as \emph{njRAT}. 
The njRAT malware family, also called Bladabindi, is a widespread remote access trojan (RAT). It allows attackers to steal passwords, log keystrokes, activate webcam and microphone, give access to the command line, and allows attackers to remotely execute and manipulate files and system registry. 

It uses the HTTP user-agent field (this is reflected in the event \emph{unexpected application} in Figure \ref{fig:njrat}) to exfiltrate sensitive information (event \emph{information stealer} in Figure \ref{fig:njrat}) from the infected machine. The communication with C\&C server uses dynamic DNS with string patterns such as \emph{maroco.dyndns.org/is-rinoy} or \emph{man2010.no-ip.org/is-ready} and specifically crafted host names. This usage of dynamic DNS is reflected in event \emph{suspicious domain from dynamic DNS} in Figure~\ref{fig:njrat}, the specific host names as event \emph{anomalous destination}. These characteristic features of njRAT are also the most important features for the transformer. 
We conclude that this explanation matches known behavior of njRAT.

\begin{figure}[th!]
	\centering
	\subfloat[ArcadeYum\label{fig:arcade_yum}]{
	 \includegraphics[width=0.49\linewidth,keepaspectratio]{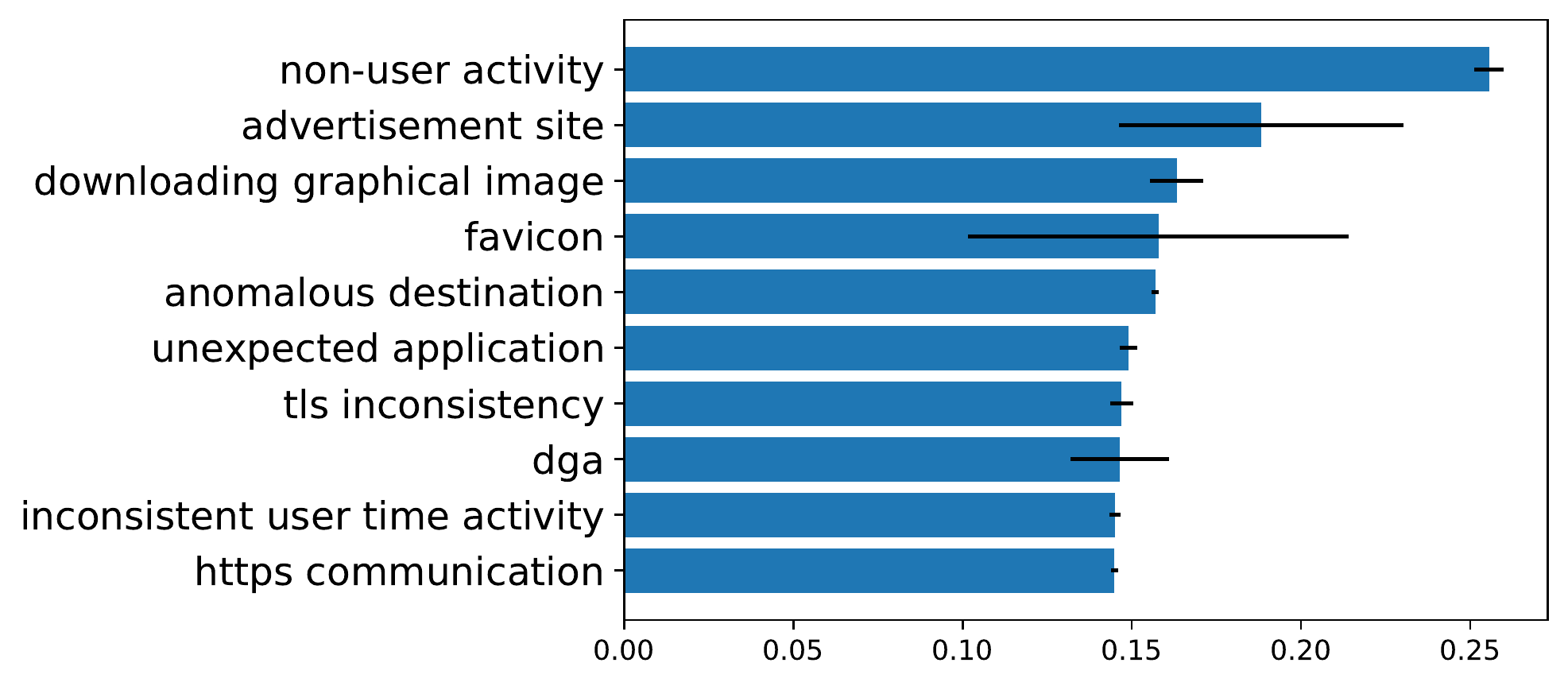}}
	 \hfil
	 \subfloat[Malicious Android firmware\label{fig:Malicious_Android firmware}]{
	 \includegraphics[width=0.49\linewidth,keepaspectratio]{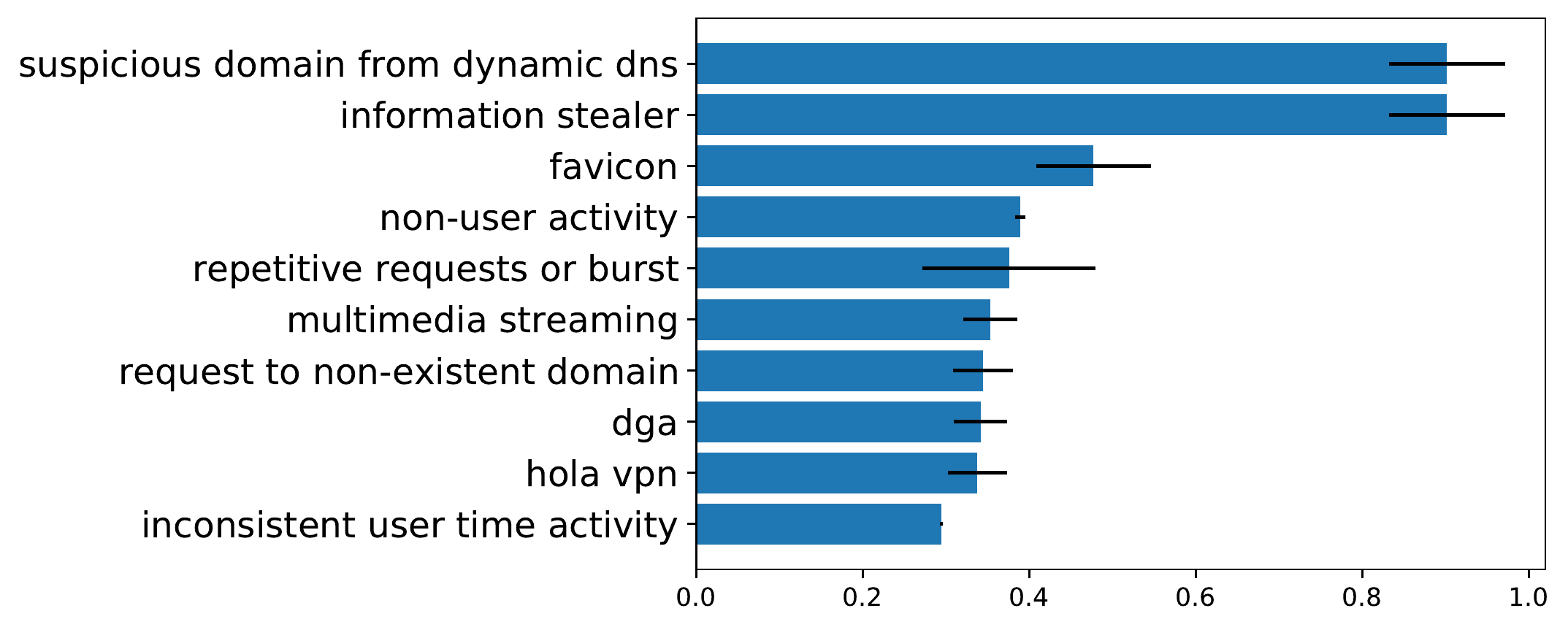}}
	 \hfil
	 \subfloat[WannaCry\label{fig:WannaCry}]{
	 \includegraphics[width=0.49\linewidth,keepaspectratio]{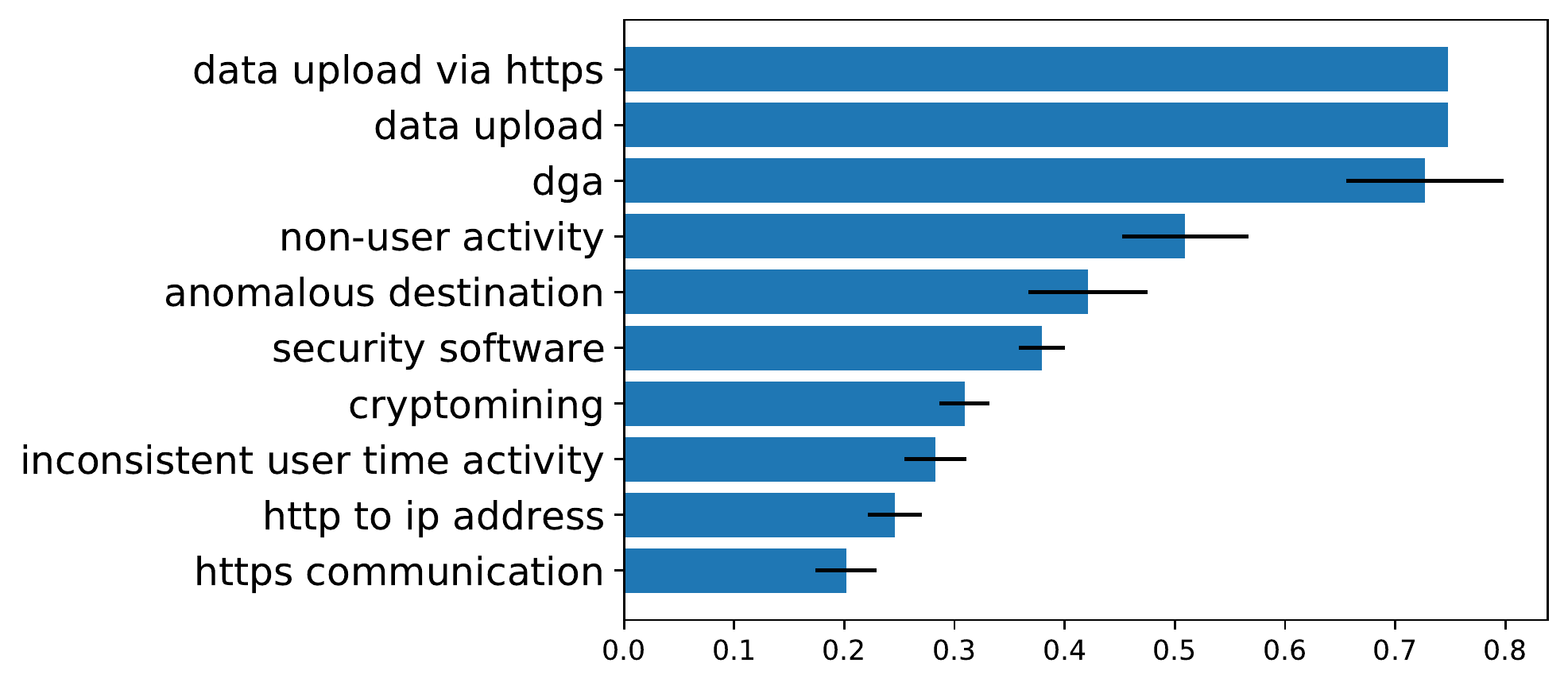}}
	 \hfil
	 \subfloat[njRAT\label{fig:njRAT}]{
	 \includegraphics[width=0.49\linewidth,keepaspectratio]{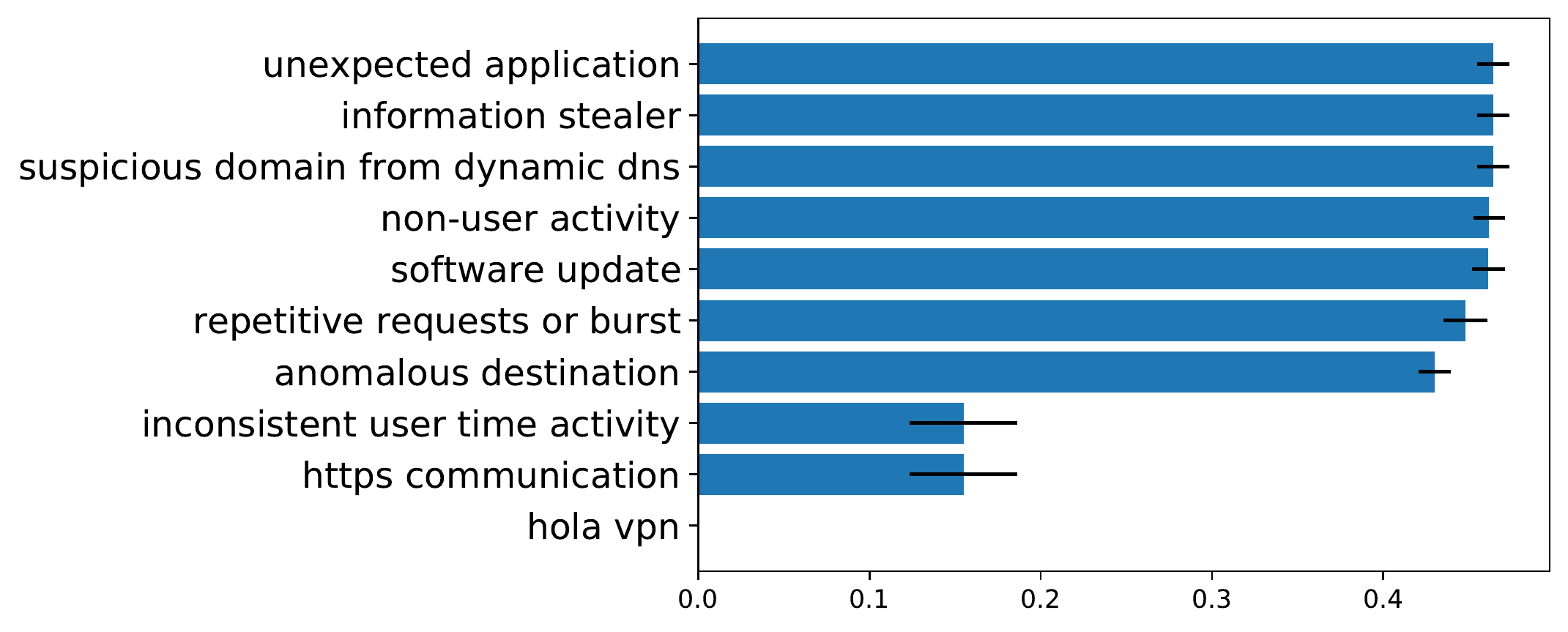}}
		\caption{Feature importances of the top 10 features for detection of different malware families. The width of the bar is computed by using the \emph{integrated gradients} method for each positively classified instance and averaging the obtained values for all network events. Error bars denote the standard deviation.}
	\label{fig:feature_importance}
\end{figure}

\subsubsection{Feature Importance}
We add the feature importance values for all the instances classified as a particular malware family. Figure~\ref{fig:feature_importance} shows the feature importance for different families. 
For \emph{njRAT}, we see that the top four features captured in Figure~\ref{fig:njRAT} matche the behavior of njRAT described above. The \emph{ArcadeYum} family is a typical example of the PUA/adware category. When installed, it starts to download large amounts of advertisement and present it as additional banners rendered on top of legitimate websites or as pop-up windows. The advertisement images are downloaded on the background without users knowledge and often from hosts that may be a source of additional infections. This behaviour is again captures by the most important features in Figure~\ref{fig:arcade_yum}. 

Most of the \emph{WannaCry} samples that we were able to detect are older versions that use DGA  domains as a kill switch---see~\cite{wannacryiocs} for details. The behavioral indicators \emph{dga, non-user activity, anomalous destination, inconsistent user time activity}  in Figure~\ref{fig:WannaCry} are related to the regular attempt to contact these DGA domains. Some of the identified samples are actually \emph{WannaMine}~\cite{wannaMine}, a crypto-mining modification of the original WannaCry malware. Their activity is captured by the \emph{cryptomining} event as well as the \emph{http to IP address}, which is the mechanism through which WannaMine downloads additional modules. 
\emph{Malicious Android firmware}, Figure~\ref{fig:Malicious_Android firmware}, is known for gathering and exfiltrating sensitive user information and using dynamic DNS to avoid blacklists. Both behaviors are represented as the top two features. The further actions depend on the type and version of the infected device. Usually, an advertisement auction service is contacted and advertisement images or videos are being displayed (\emph{multimedia streaming, repetitive requests, non-user activity, dga}).

\section{Conclusion}
We have studied the problem of identifying threats based on sequences of sets of human-comprehensible network events that are the output of a wide array of specialized detectors. We can conclude that the \emph{transformer} architecture outperforms both the CNN and LSTM models at identifying threat IDs, malware families, and malware categories. Furthermore, unsupervised pre-training improves the transformer's performance over supervised learning from scratch. We use the \emph{integrated gradients} method to determine the sequence of the most important network events that constitute \emph{indicators of compromise} which can be verified by security analysts. Our detailed analysis of the njRAT malware shows that the sequence of highly important events corresponds to the known behavior of the virus. We can conclude that for the four most frequent malware families, the network events that reach the highest aggregated feature importance across all occurrences match known indicators of compromise.

\appendix
\section{Appendix}
\label{sec:appendix}

This section reports on detailed analysis results by measuring precision-recall curves for the detection of threat IDs, malware families, and malware categories. Precision---the fraction of alarms that are not false alarms---directly measures the amount of unnecessary workload imposed on security analysts, while recall quantifies the detection rate. We also compare the models in terms of ROC cuves because these curves are invariant to class ratios.

\begin{figure}[!th]
    \centering    
    \subfloat[ROC curves (log-scale for FPR)\label{fig:cluster_label_roc}]{
        \includegraphics[width=0.49\textwidth,keepaspectratio]{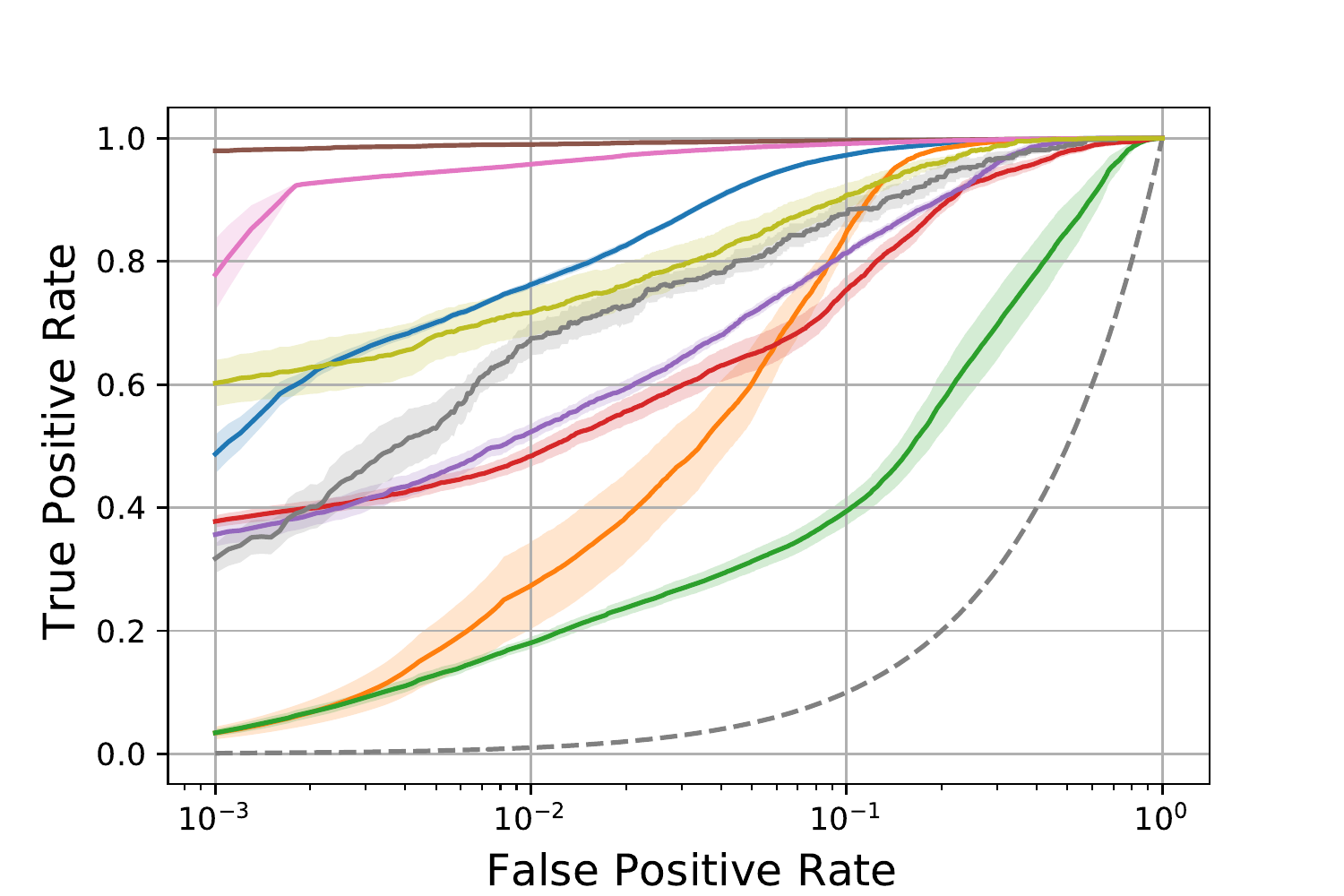}}
    \hfil
    \subfloat[Precision-recall curves\label{fig:cluster_label_pr}]{
        \includegraphics[width=0.49\textwidth,keepaspectratio]{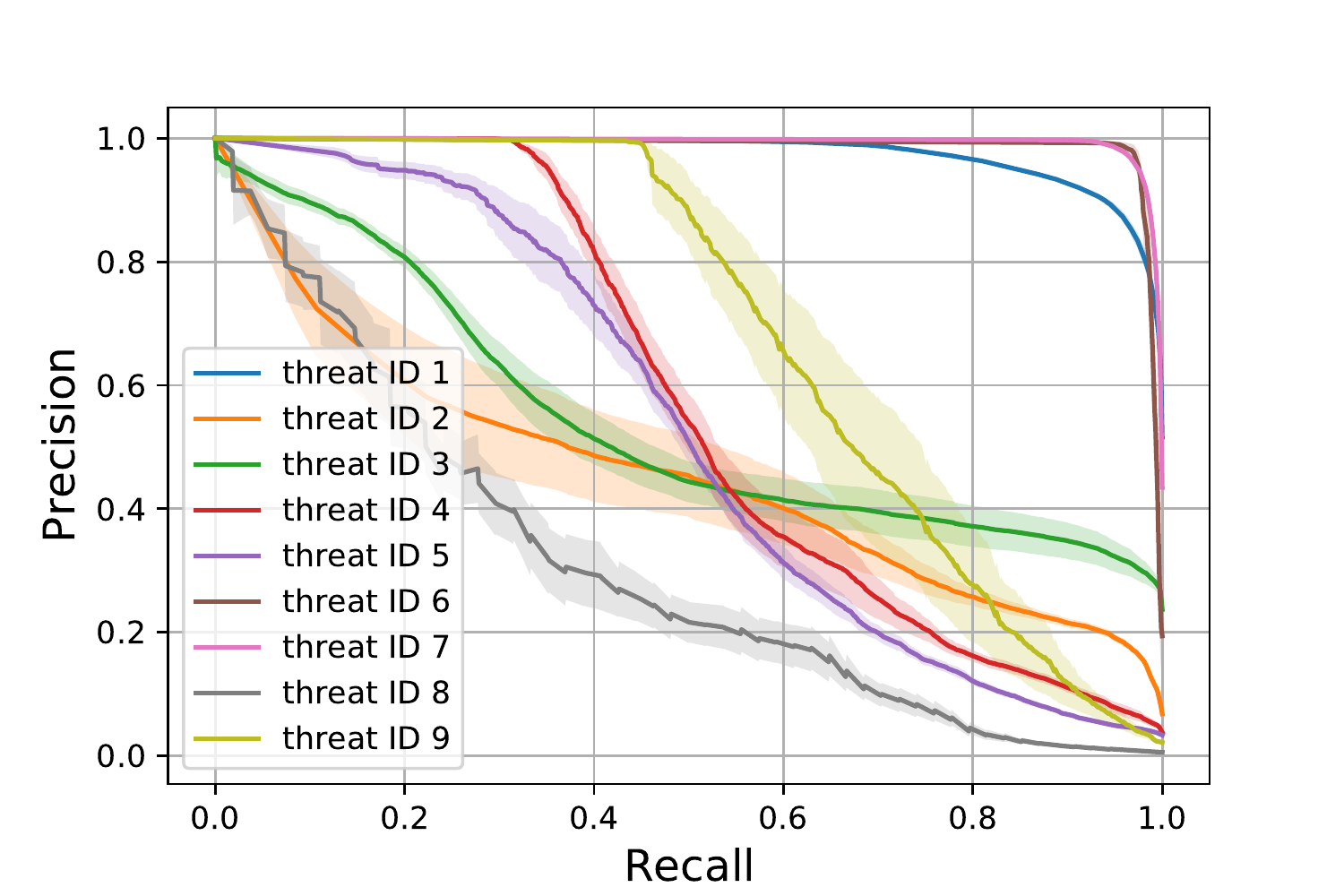}}
    
    \caption{ROC and precision-recall curves for the detection of threat IDs on the evaluation data. The dashed line shows the performance for random guessing and the colored bands the standard error.}
	\label{fig:cluster_label}
\end{figure}

\subsection{Threat ID Evaluation}
Figure~\ref{fig:cluster_label} shows the ROC and the precision-recall curves for detecting different threats using the transformer model. As Figure~\ref{fig:cluster_label_roc} shows, threat IDs that have a one-to-one relationship with a malware category (threat IDs 6, 7, 8, 9) can be detected more easily than threat IDs that share the same malware category with several other threat IDs. We see that the threat IDs 2, 3, and 4 that belong to the category of \emph{potentially unwanted application} are much harder to detect than other threat IDs. We can be explained by the similar behavior of multiple threat IDs within a category. In total, the transformer is able to detect 6 out of 9 threat IDs with a precision of 80\% and a recall of at approximately 40\%. 

\begin{figure}[!ht]
    \centering    
    \subfloat[ROC curves (log-scale for FPR)\label{fig:malware_categories_roc}]{
        \includegraphics[width=0.49\textwidth,keepaspectratio]{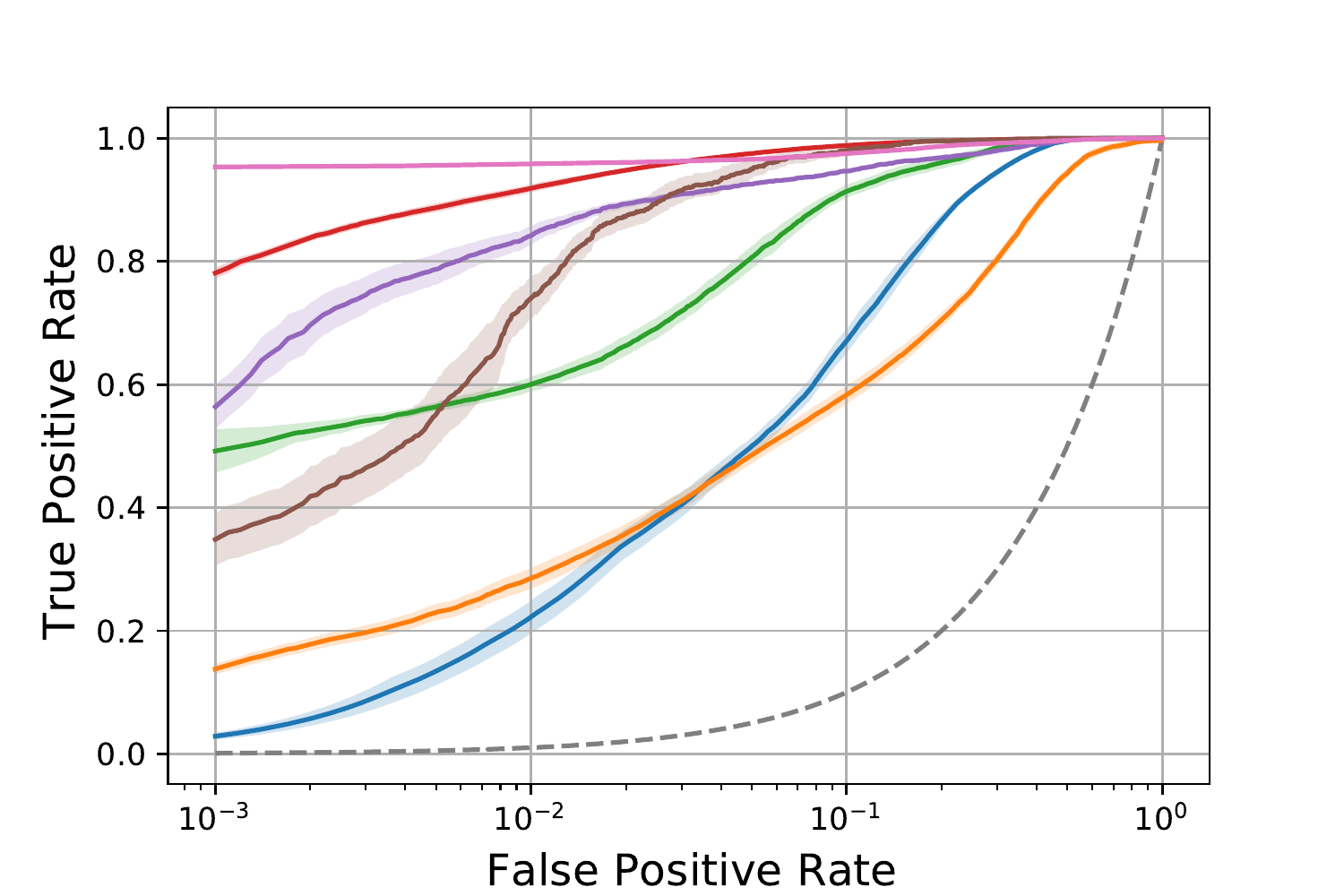}}
    \hfil
    \subfloat[Precision-recall curves\label{fig:malware_categories_pr}]{
        \includegraphics[width=0.49\textwidth,keepaspectratio]{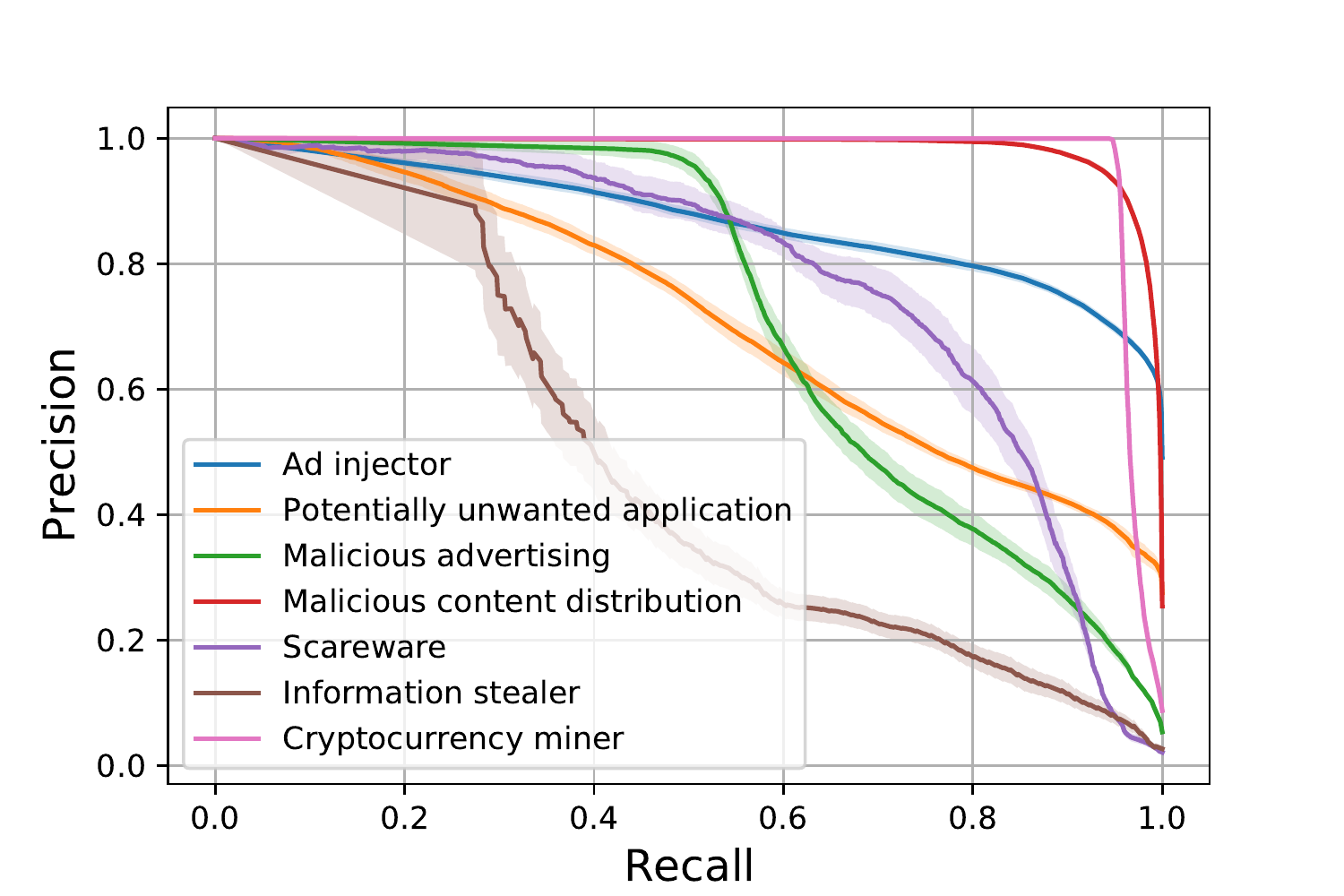}}
    
    \caption{ROC and precision-recall curves for the detection of malware categories on the evaluation data. The dashed line shows the performance for random guessing and the colored bands the standard error.}
	\label{fig:malware_categories}
\end{figure}

\subsection{Malware-Category Evaluation}
Figure~\ref{fig:malware_categories} shows the ROC and the precision-recall curves for detecting different malware categories using the transformer model. The high-prevalence malware categories potentially \emph{unwanted application} and \emph{ad injector} are much harder to detect than the rare malware categories \emph{cryptocurrency miner} and \emph{malicious content distribution}. This can be explained by the larger behavioral variations in the frequent categories.Regarding the precision-recall curves, we can conclude that the transformer is able to detect 6 out of 7 malware categories with a precision of 80\% with a recall higher than 40\%. 

\begin{figure}[!ht]
    \centering
    \subfloat[ROC curves (log-scale for FPR)\label{fig:malware_fam_roc}]{
        \includegraphics[width=0.49\textwidth,keepaspectratio]{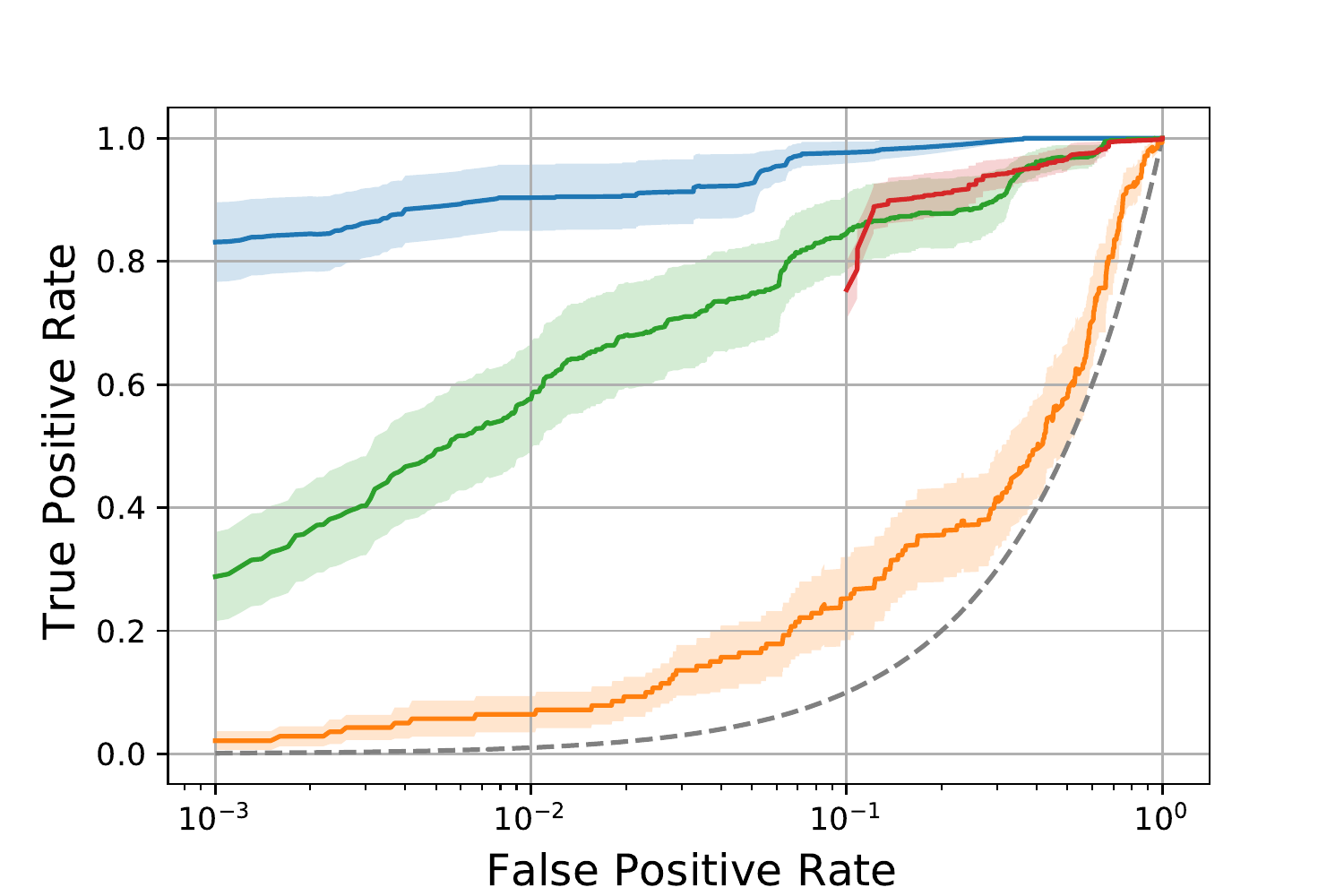}}
    \hfil
    \subfloat[Precision-recall curves\label{fig:malware_fam_pr}]{
        \includegraphics[width=0.49\textwidth,keepaspectratio]{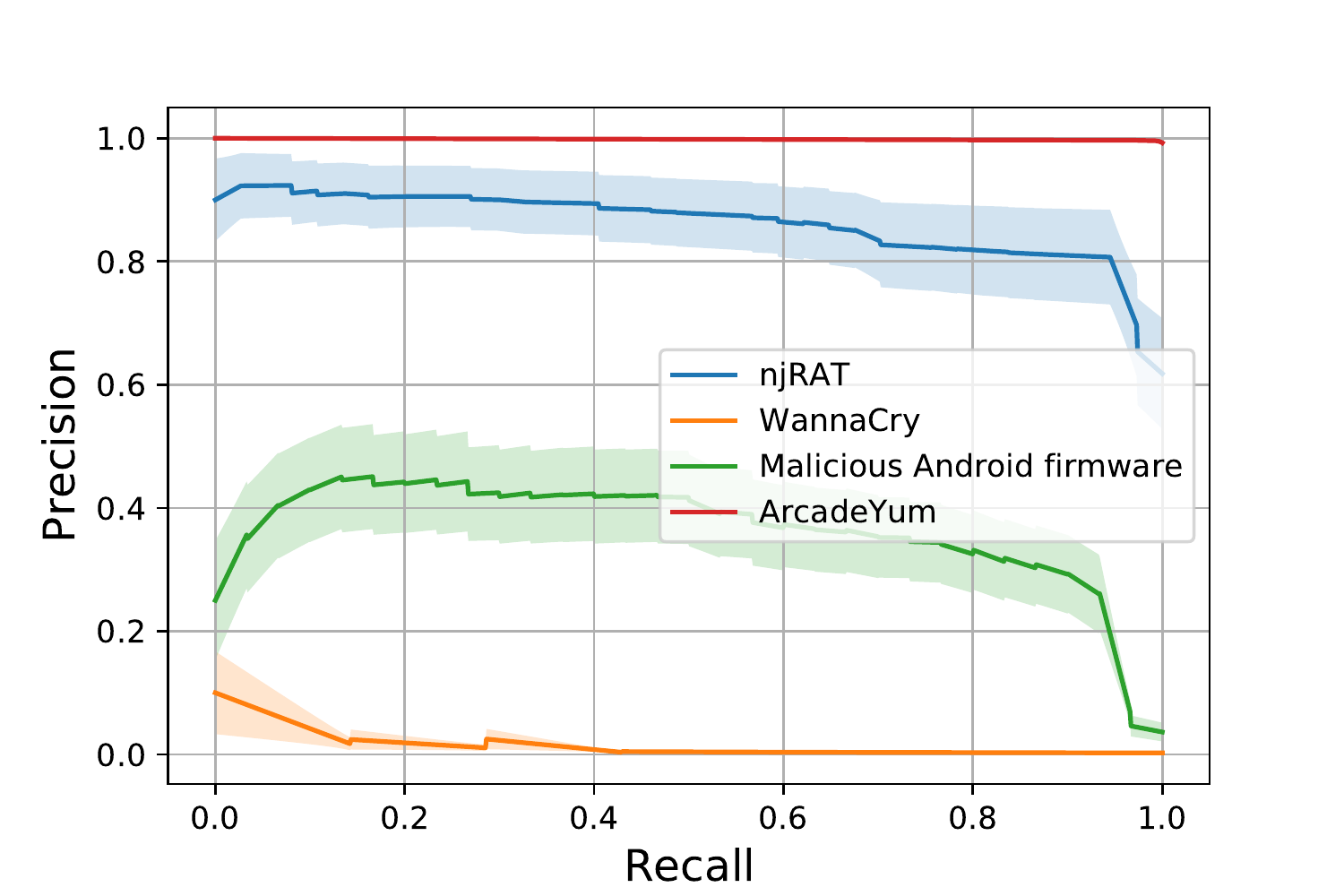}}
    
    \caption{Malware Family Evaluation. Performance for models on test set. The dashed line shows the performance for random guessing and the colored bands the standard error.}
	\label{fig:malware_fam}
\end{figure}

\subsection{Malware Family Evaluation}
Figure~\ref{fig:malware_fam} shows the ROC and the precision-recall curves for detecting different malware families using the transformer model. Because of the highly unbalanced class ratios, attention should be given to the ROC curves of Figure~\ref{fig:malware_fam_roc}; the precision-recall curves of  Figure~\ref{fig:malware_fam_pr} are less informative, but are included  for completeness. We can conclude that the transformer performs best on the two information stealers. Because we only observe under 100 instances not belonging to the malware family ArcadeYum, we can only draw the ROC curve up to a FPR of 0.1. In general, we see that the transformer is able to distinguish between malware families. Only the detection of WannaCry is significantly worse; this finding is plausible because especially newer versions of WannaCry create minimal network traffic. 

\bibliographystyle{splncs04}
\bibliography{preprint}

\end{document}